\documentclass{article} 
\usepackage{iclr2016_workshop,times}
\usepackage{hyperref}
\usepackage{url}
\usepackage{amsmath}
\usepackage{caption}
\usepackage{subcaption}
\usepackage[pdftex]{graphicx}

\title{Resnet in Resnet: Generalizing Residual \\ Architectures}

\author{Sasha Targ\thanks{Equal contribution. Author ordering determined by coin flip.}, Diogo Almeida\footnotemark[1], Kevin Lyman \\
Enlitic\\
\texttt{sasha.targ@ucsf.edu,\{diogo,kevin\}@enlitic.com} \\
}

\begin{document}
\maketitle

\begin{abstract}
Residual networks (ResNets) have recently achieved state-of-the-art on challenging computer vision tasks. We introduce Resnet in Resnet (RiR): a deep dual-stream architecture that generalizes ResNets and standard CNNs and is easily implemented with no computational overhead. RiR consistently improves performance over ResNets, outperforms architectures with similar amounts of augmentation on CIFAR-10, and establishes a new state-of-the-art on CIFAR-100.
\end{abstract}

\section{Introduction}
Recently proposed residual networks (ResNets) get state-of-the-art performance on the ILSVRC 2015 classification task \citep{imagenet} and allow training of extremely deep networks up to more than 1000 layers \citep{resnet}. Similar to highway networks, residual networks make use of identity shortcut connections that enable flow of information across layers without attenuation that would be caused by multiple stacked non-linear transformations, resulting in improved optimization \citep{highway}. In residual networks, shortcut connections are not gated and untransformed input is always transmitted. While the empirical performance of ResNets in \cite{resnet} is very impressive, current residual network architectures have several potential limitations: identity connections as implemented in the current ResNet leads to accumulation of a mix of levels of feature representations at each layer, even though in a deep network some features learned by earlier layers may no longer provide useful information in later layers \citep{forget}. \\

A hypothesis of the ResNet architecture is that learning identity weights is difficult, but by the same argument, it is difficult to learn the additive inverse of identity weights needed to remove information from the representation at any given layer. The fixed size layer structure of the residual block modules also enforces that residuals must be learned by shallow subnetworks, despite evidence that deeper networks are more expressive \citep{expressivity}. We introduce a generalized residual architecture that combines residual networks and standard convolutional networks in parallel residual and non-residual streams. We show architectures using generalized residual blocks retain the optimization benefits of identity shortcut connections while improving expressivity and the ease of removing unneeded information. We then present a novel architecture, ResNet in ResNet (RiR), which incorporates these generalized residual blocks within the framework of a ResNet and demonstrate state-of-the-art performance of the RiR architecture on CIFAR-100.

\section{Generalizing Residual Network Architectures}
The modular unit of the generalized residual network architecture is a generalized residual block consisting of parallel states for a residual stream, $\mathbf{r}$, which contains identity shortcut connections and is similar to the structure of a residual block from the original ResNet with a single convolutional layer (parameters $W_{l, r \to r}$), and a transient stream, $\mathbf{t}$, which is a standard convolutional layer ($W_{l, t \to t}$). Two additional sets of convolutional filters in each block ($W_{l, r \to t}$, $W_{l, t \to r}$) also transfer information across streams.
\begin{align*}
  \mathbf{r}_{l+1} &= \sigma(\mbox{conv}(\mathbf{r}_l, W_{l, r \to r}) + \mbox{conv}(\mathbf{t}_l, W_{l, t \to r}) + \mbox{shortcut}(\mathbf{r}_l)) \addtocounter{equation}{1}\tag{\theequation} \label{eqn} \\
  \mathbf{t}_{l+1} &= \sigma(\mbox{conv}(\mathbf{r}_l, W_{l, r \to t}) + \mbox{conv}(\mathbf{t}_l, W_{l, t \to r}))
\end{align*}
Same-stream and cross-stream activations are summed (along with the shortcut connection for the residual stream) before applying batch normalization and ReLU nonlinearities (together $\sigma$) to get the output states of the block (Equation \ref{eqn}) \citep{bn}. The function of the residual stream $\mathbf{r}$ resembles that of the original structure of the ResNet \citep{resnet} with shortcut connections between each unit of processing, while the transient stream $\mathbf{t}$ adds the ability to process information from either stream in a nonlinear manner without shortcut connections, allowing information from earlier states to be discarded. The form of the shortcut connection can be an identity function with the appropriate padding or a projection as in \cite{resnet}. We implement the generalized residual block as a single convolutional layer with a modified initialization, which we call ResNet Init (see Appendix \ref{subsection:implementation} for detail).
\begin{figure}[!ht]
  \centering
  \subcaptionbox{\label{fig:resnet}}{\includegraphics[width=0.9in]{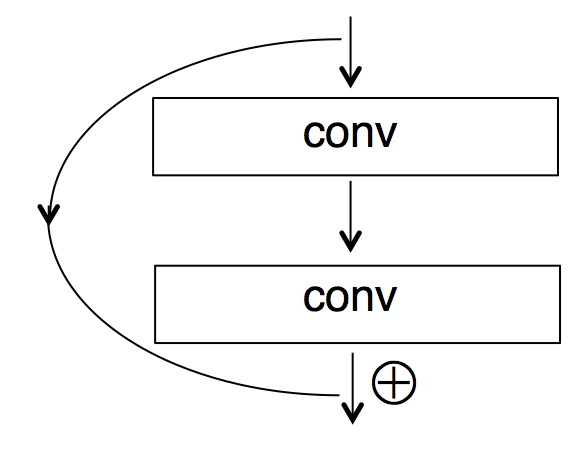}}\hspace{1.0em}%
  \subcaptionbox{\label{fig:resnet-init}}{\includegraphics[width=1.55in]{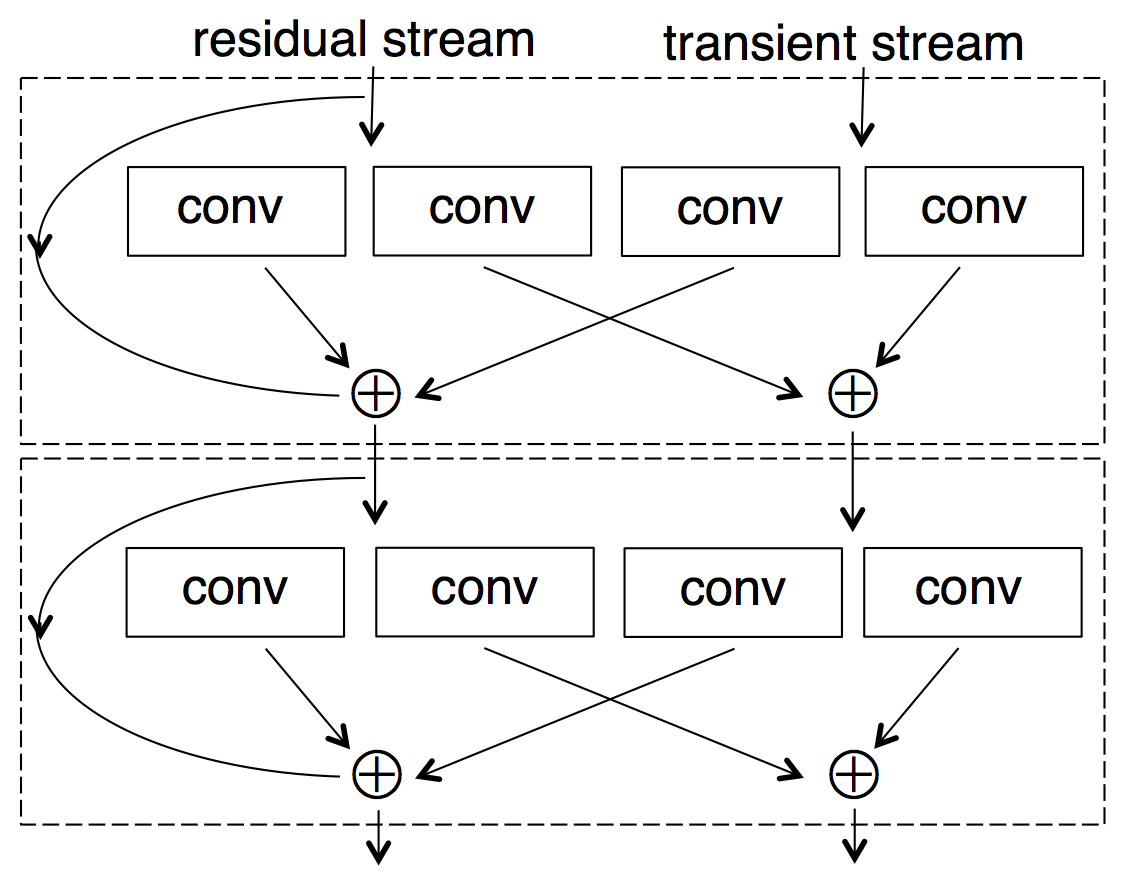}}\hspace{1.0em}%
  \subcaptionbox{\label{fig:resnet-init-to-block}}{\includegraphics[width=1.65in]{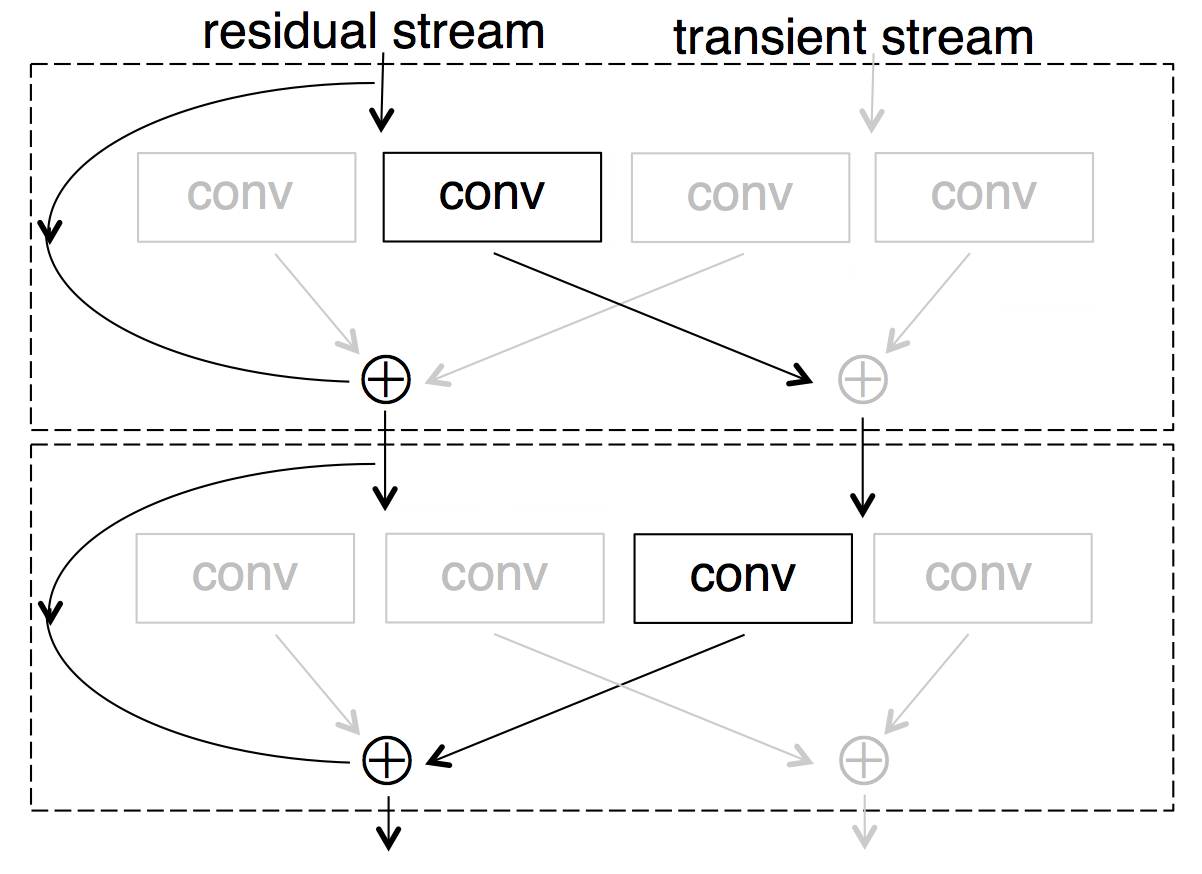}}\hspace{1.0em}%
  \subcaptionbox{\label{fig:RiR}}{\includegraphics[width=0.9in]{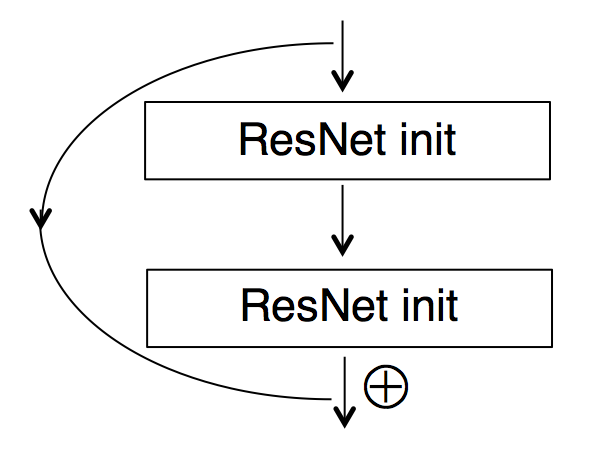}}
  \caption{(a) 2-layer ResNet block. (b) 2 generalized residual blocks (ResNet Init). (c) 2-layer ResNet block from 2 generalized residual blocks (grayed out connections are 0). (d) 2-layer RiR block.}
  \label{fig:resnets}
\end{figure}

\begin{figure}[!ht]
  \centering
    \includegraphics[scale=0.3]{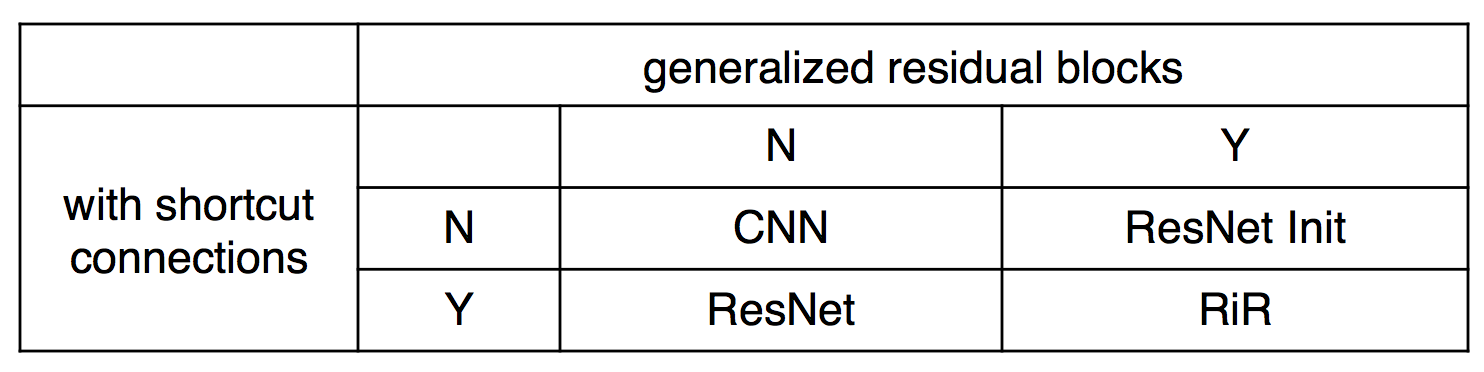}
    \caption{Relationship between standard CNN, ResNet, ResNet Init, and RiR architectures.}
  \label{fig:relationship-table}
\end{figure}

The generalized residual block can act as either a standard CNN layer (by learning to zero the residual stream) or a single-layer ResNet block (by learning to zero the transient stream). By repeating the generalized residual block several times, the generalized residual architecture has the expressivity to learn anything in between, including the standard 2-layer ResNet block (Figure \ref{fig:resnet-init-to-block}). This architecture allows the network to learn residuals with a variable effective number of processing steps before addition back into the residual stream, which we investigate by visualizations. The generalized residual block is not specific to CNNs, and can be applied to standard fully connected layers and other feedforward layers. Replacing each of the convolutional layers within a residual block from the original ResNet (Figure \ref{fig:resnet}) with a generalized residual block (Figure \ref{fig:resnet-init}) leads us to a new architecture we call ResNet in ResNet (RiR) (Figure \ref{fig:RiR}). In Figure \ref{fig:relationship-table}, we summarize the relationship between standard CNN, ResNet Init, ResNet, and RiR architectures.

\section{Experiments}
\begin{table}[t]
\begin{minipage}[t]{.5\linewidth}

\caption{Comparison of our architecture with state-of-the-art architectures on CIFAR-10.}
\label{wide-cifar10}
\begin{center}
\begin{tabular}{ll}
\multicolumn{1}{c}{\bf Model}  &\multicolumn{1}{c}{\bf Accuracy (\%)}
\\ \hline
Highway Network  & 92.40 \\
ResNet (32 layers)  & 92.49 \\
ResNet (110 layers)  & 93.57 \\
Large ALL-CNN  & 95.59 \\
{\bf Fractional Max-Pooling}  & {\bf 96.53} \\
\hline
18-layer + wide CNN & 93.64 \\
18-layer + wide ResNet & 93.95  \\
18-layer + wide ResNet Init & 94.28 \\
18-layer + wide RiR & 94.99
\end{tabular}
\end{center}
\end{minipage}
\hspace{0.1cm}%
\begin{minipage}[t]{.5\linewidth}
\caption{Comparison of our architecture with state-of-the-art architectures on CIFAR-100.}
\label{wide-cifar100}
\begin{center}
\begin{tabular}{ll}
\multicolumn{1}{c}{\bf Model}  &\multicolumn{1}{c}{\bf Accuracy (\%)}
\\ \hline
Highway Network & 67.76 \\
ELU-Network  & 75.72 \\
\hline
18-layer + wide CNN & 75.17 \\
18-layer  + wide ResNet & 76.58  \\
18-layer + wide ResNet Init & 75.99 \\
{\bf 18-layer + wide RiR} & {\bf 77.10}
\end{tabular}
\end{center}
\end{minipage}%
\end{table}


We evaluate our architectures on CIFAR-10 and CIFAR-100 datasets \citep{cifar-data} and report the best results found after a grid search on hyperparameters including learning rate, L2 penalty, initialization among Xavier \citep{xavier-init}, MSR \citep{msr-init}, and orthogonal \citep{orthogonal-init}, optimizer among SGD with momentum, SGD with Nesterov momentum \citep{nesterov}, Adam \citep{adam}, and RMSProp \citep{rmsprop}, and the type of shortcut connections in the residual blocks. We optimize the hyperparameters for the original ResNet architecture and use SGD with momentum of 0.9, a minibatch size of 500, L2 penalty of 0.0001, and train for 82 epochs. The learning rate was scaled by 0.1 after epochs 42 and 62 and MSR initialization was used for all weight tensors. Test time batch normalization statistics were approximated using an exponential moving average of training batch normalization statistics. A projection with a 3x3 convolution was used for residual blocks that increase dimensionality, and all other shortcut connections were identity. We use equal numbers of filters for the residual and transient streams of the generalized residual network, but optimizing this hyperparameter could lead to further potential improvements.

\begin{figure}[ht]
\centering
\begin{minipage}[b]{0.45\linewidth}
\includegraphics[scale=0.45]{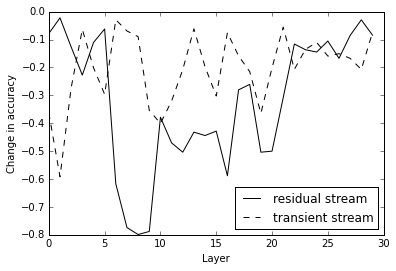}
\caption{Effect of ablating each stream of the generalized residual network architecture}
\label{fig:ablation}
\end{minipage}
\quad
\begin{minipage}[b]{0.45\linewidth}
\includegraphics[scale=0.33]{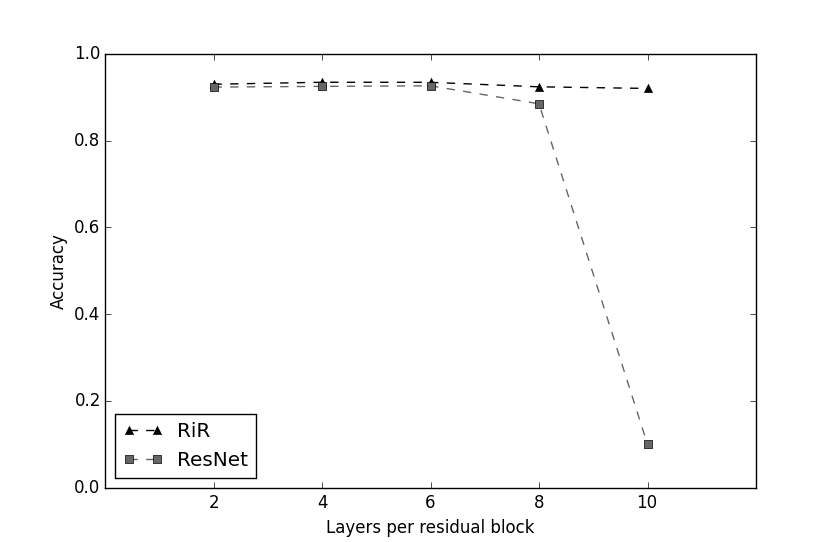}
\caption{ResNet and RiR with increased layers/block. All models have 15 blocks.}
\label{fig:more_layers}
\end{minipage}
\end{figure}

In our experiments, the ResNet Init architecture shows consistent improvement over standard CNN architectures, and the RiR architecture outperforms the original ResNet (Table \ref{original-cifar10}). We find the RiR architecture performs well across a range of numbers of blocks and layers in each block and that ResNet Init applied to existing architectures, such as ALL-CNN-C \citep{allconv}, yields improvement over standard initialization (Tables \ref{allconv-cifar10},  \ref{combo-cifar10}). Because each stream uses only half the total filters, we investigate the effect of our architectures on a wider 18-layer network (Tables \ref{wide-cifar10}, \ref{wide-cifar100}). We find this RiR architecture is remarkably effective, obtaining competitive results on CIFAR-10 with only standard augmentation by random crops and horizontal flips, and state-of-the-art results on CIFAR-100. We visualize the effect of zeroing learned connections of each stream in a trained ResNet Init model a single layer at a time, which shows both streams contribute to accuracy and relative use of residual and transient streams changes at different stages of processing (Figure \ref{fig:ablation}). In Figure \ref{fig:more_layers}, we show performance of the RiR architecture is robust to increasing depth of residual blocks and the RiR architecture allows training of deeper residuals compared to the original ResNet.

\section{Related Work}
Interacting transformation streams in which only one stream includes shortcut connections are also used in blocks of LSTM and Grid-LSTM networks \citep{lstm, grid-lstm}. However, in contrast to highway networks which control flow through shortcut connections via input-dependent carry and transform gates, and to memory and hidden states of LSTM and Grid-LSTM blocks, flow of information between the residual and transient states of the generalized residual block does not use gates and can thus be implemented with no additional parameters over a standard feedforward network \citep{highway}. Another difference between previous architectures and the generalized residual block we present is that while memory ($\mathbf{m}$) and hidden ($\mathbf{h}$) states in an LSTM or Grid-LSTM block are calculated sequentially (with $\mathbf{h}_l = \mathbf{o}_l \odot \mbox{tanh}(\mathbf{m}_l)$), transformation of residual and memory streams of the generalized residual block occurs in parallel and depends only on the learned convolutional filters at each layer without further constraints on their relation. The SCRN architecture of \cite{scrn} also uses hidden and context units together within a single layer to learn longer term information, which behave similarly to the transient and residual streams, but SCRN only allows unidirectional flow from context to hidden units and connections between context units are fixed, in contrast to bidirectional flow between streams and learned connections for both transient and residual streams in our generalized residual architecture.

\section{Conclusion}
We present a generalized residual architecture which be simply implemented by a modified initialization scheme, ResNet Init, and apply it to the original ResNet to create a novel RiR architecture which achieves state-of-the-art results. Future work includes additional study of RiR and related residual models to further determine the cause of their beneficial effects.

\bibliography{iclr2016_workshop}
\bibliographystyle{iclr2016_workshop}

\newpage
\section{Appendix}

\begin{table}[!ht]
\begin{minipage}[t]{.5\linewidth}
\caption{Test set accuracy of baseline 32-layer CNN \citep{resnet} on CIFAR-10.}
\label{original-cifar10}
\begin{center}
\begin{tabular}{ll}
\multicolumn{1}{c}{\bf Model}  &\multicolumn{1}{c}{\bf Accuracy (\%)}
\\ \hline
ResNet~\citep{resnet} & 92.49 \\
\hline
CNN & 89.03 \\
ResNet & 92.32  \\
ResNet Init & 89.62 \\
{\bf RiR} & {\bf 92.97}
\end{tabular}
\end{center}
\end{minipage}
\hspace{0.1cm}%
\begin{minipage}[t]{.5\linewidth}
\caption{Performance of ALL-CNN-C from \cite{allconv} with batch normalization \citep{bn}. The original model without batch normalization had 90.92\% accuracy.}
\label{allconv-cifar10}
\begin{center}
\begin{tabular}{ll}
\multicolumn{1}{c}{\bf Model}  &\multicolumn{1}{c}{\bf Accuracy (\%)}
\\ \hline
Standard initialization & 93.22 \\
{\bf ResNet Init} & {\bf 93.42} \\
\end{tabular}
\end{center}
\end{minipage}
\end{table}

\subsection{Generalized Residual Block Implementation}
\label{subsection:implementation}

We implement the generalized residual block with a modified initialization of a standard convolutional or fully connected (FC) layer that combines the identity shortcut with the desired linear transformation (convolution or matrix multiplication), which we call ResNet Init, and concatenating tensors for the residual $\mathbf{r}$ and transient $\mathbf{t}$ streams to form a single tensor $\mathbf{x}$. Because the identity shortcut and the results of the same-stream and cross-stream transformations are summed to give the output for each stream, linear operations on $\mathbf{r}$ and $\mathbf{t}$ can be composed into a single linear operation on $\mathbf{x}$ (see Equation \ref{init} for an example of ResNet Init applied to an FC layer). 

\begin{align*}
\mathbf{x}_{l+l} = \sigma(W'_l \mathbf{x}_l) \Leftrightarrow
\left[ \begin{array}{c} \mathbf{r}_{l+1} \\ \mathbf{t}_{l+1} \end{array} \right] = \sigma((\begin{bmatrix} W_{l,r \to r} & W_{l,t \to r} \\ W_{l,r \to t} & W_{l,t \to t} \end{bmatrix} + \begin{bmatrix} I & 0 \\ 0 & 0 \end{bmatrix}) \times \left[ \begin{array}{c} \mathbf{r}_{l} \\ \mathbf{t}_{l} \end{array} \right]) \addtocounter{equation}{1}\tag{\theequation} \label{init}
\end{align*}

To implement ResNet Init in a single FC layer, we concatenate weight matrices initialized by any existing scheme and then add a partial identity matrix (with 1s only on the first half of the diagonal) to the concatenated weight matrix. To implement ResNet Init in a single convolutional layer, we first concatenate convolutional kernels along the input dimension ($W_{l,x \to r} = W_{l,r \to r} + W_{l,t \to r}$ and $W_{l,x \to t} = W_{l,r \to t} + W_{l,t \to t}$) and the filter dimension ($W_{l, x \to x} = W_{l,x \to r} + W_{l,x \to t}$), and then similarly add half of an identity kernel. The output of the generalized residual block implemented as a standard layer with modified initialization ($W'_l$) is exactly equivalent to the output if it were implemented as separate linear operations ($W_{l,r \to r}, W_{l,t \to r}, W_{l,r \to t}, W_{l,t \to t},$ and $I$). 

The generalized residual block implemented using modified initialization will perform differently from one implemented as separate linear operations when weight regularization that pulls all weights towards zero is present, such as L2 regularization. To maintain equivalence of implementation, we subtract the partial identity from the weights prior to application of weight decay.
\hspace{0.1cm}%



\begin{table}
\begin{minipage}[t]{.5\linewidth}
\caption{Accuracy of RiR with different numbers of blocks and layers per block on CIFAR-10.}
\label{combo-cifar10}
\begin{center}
\begin{tabular}{lll}
\multicolumn{1}{c}{\bf Blocks}  &\multicolumn{1}{c}{\bf Layers/Block} &\multicolumn{1}{c}{\bf Accuracy (\%)}
\\ \hline
15 & 2 & 92.87 \\
3 & 10 & 90.06 \\
6 & 5 & 92.98 \\
15 & 5 & 92.11 \\
{\bf 9} & {\bf 3} & {\bf 93.23} \\
\end{tabular}
\end{center}
\end{minipage}
\hspace{0.1cm}%
\begin{minipage}[t]{.5\linewidth}
\caption{Accuracy of RiR and ResNet with different numbers of layers per block on CIFAR-10.}
\label{more-layers-table}
\begin{center}
\begin{tabular}{lll}
\multicolumn{1}{c}{\bf Layers/Block} &\multicolumn{1}{c}{\bf ResNet (\%)} &\multicolumn{1}{c}{\bf RiR (\%)}
\\ \hline
2 & 92.32 & 92.97 \\
4 & 92.48 & {\bf 93.43} \\
6 & 92.61 & 93.42 \\
8 & 88.47 & 92.38 \\
10 & 10.00 & 92.01 \\
\end{tabular}
\end{center}
\end{minipage}
\end{table}

\newpage
\subsection{Architectures}
\begin{table}[!ht]
\caption{Description of network architectures used in experiments comparing performance of standard CNN, ResNet, ResNet Init, and RiR models. In standard CNN and ResNet Init models, identity connections of the residual blocks listed below are set to 0. For CIFAR-10 models, the output layer has 10 units and for CIFAR-100 models, the output layer has 100 units. When not otherwise specified, convolutions are performed with stride 1.}
\label{original-arch}
\begin{center}
\begin{tabular}{ll}
\multicolumn{1}{c}{Baseline 32-layer CNN architecture \citep{resnet}}
\\ \hline
3x3 conv., 16 filters, BN, ReLU \\
Residual block: 2x 3x3 conv., 16 filters, BN, ReLU \\
Residual block: 2x 3x3 conv., 16 filters, BN, ReLU \\
Residual block: 2x 3x3 conv., 16 filters, BN, ReLU \\
Residual block: 2x 3x3 conv., 16 filters, BN, ReLU \\
Residual block: 2x 3x3 conv., 16 filters, BN, ReLU \\
Residual block: 2x 3x3 conv. (first with stride 2), 32 filters, BN, ReLU \\
Residual block: 2x 3x3 conv., 32 filters, BN, ReLU \\
Residual block: 2x 3x3 conv., 32 filters, BN, ReLU \\
Residual block: 2x 3x3 conv., 32 filters, BN, ReLU \\
Residual block: 2x 3x3 conv., 32 filters, BN, ReLU \\
Residual block: 2x 3x3 conv. (first with stride 2), 64 filters, BN, ReLU \\
Residual block: 2x 3x3 conv., 64 filters, BN, ReLU \\
Residual block: 2x 3x3 conv., 64 filters, BN, ReLU \\
Residual block: 2x 3x3 conv., 64 filters, BN, ReLU \\
Residual block: 2x 3x3 conv., 64 filters, BN, ReLU \\
Global mean pool \\
FC layer, 10 or 100 units \\ 
Softmax \\
Total number of parameters: 0.46M (CNN and ResNet Init) / 0.49M (ResNet and RiR)
\\
\\
\multicolumn{1}{c}{18-layer and wide CNN architecture}
\\ \hline
3x3 conv., 96 filters, BN, ReLU \\
Residual block: 2x 3x3 conv., 96 filters, BN, ReLU \\
Residual block: 2x 3x3 conv., 96 filters, BN, ReLU \\
Residual block: 2x 3x3 conv. (first with stride 2), 192 filters, BN, ReLU \\
Residual block: 2x 3x3 conv., 192 filters, BN, ReLU \\
Residual block: 2x 3x3 conv., 192 filters, BN, ReLU \\
Residual block: 2x 3x3 conv. (first with stride 2), 384 filters, BN, ReLU \\
Residual block: 2x 3x3 conv., 384 filters, BN, ReLU \\
Residual block: 2x 3x3 conv., 384 filters, BN, ReLU \\
1x1 conv., 10 or 100 filters \\
Global mean pool \\
Softmax \\
Total number of parameters: 9.5M (CNN and ResNet Init) / 10.3M (ResNet and RiR)
\end{tabular}
\end{center}
\end{table}

\end{document}